# 如何證明粒子群最佳化演算法超參數的最佳值？

Abel C. H. Chen
Chunghwa Telecom Co., Ltd.
chchen.scholar@gmail.com; ORCID 0000-0003-3628-3033

**摘要**

近年來，許多群智能最佳化演算法陸續被提出，並且被應用來解決各種最佳化問題。然而，群智能最佳化演算法可能有許多超參數值需要設定。例如，粒子群最佳化演算法雖然廣泛被應用，並且有較高的最佳化效能，但需要設定慣性速度權重、個體最佳解權重、群體最佳解權重等多個超參數值。因此，本研究提出一個分析框架，運用數學模型分析粒子群最佳化演算法在不同的適應函數中的平均適應函數值，並且從最小化平均適應函數值來決定最合適的超參數值。由實驗結果顯示，採用本研究提出的超參數值，可以有較高的收斂速度和較小的適應函數值。

**關鍵字**：粒子群最佳化演算法、超參數、最佳值。

## 1. 前言

為解決各式各樣的最佳化問題，許多群智能最佳化演算法被提出，包含有粒子群最佳化演算法(Particle Swarm Optimization, PSO)[1]、海豚群演算法(Dolphin Swarm Algorithm, DSA)[2]、灰狼最佳化演算法(Grey Wolf Optimizer, GWO)[3]、基因演算法(Genetic Algorithm, GA)[4]、人工蜂群演算法(Artificial Bee Colony, ABC)[5]、鼠群最佳化演算法(Rat Swarm Optimizer, RSO)[6]等。然而，在不同的群智能最佳化演算法可能有不同的超參數需要設定，並且將可能影響收斂速度和最佳化結果。因此，如何設置群智能最佳化演算法的超參數是一個重要的研究議題。

其中，粒子群最佳化演算法是一種被廣泛應用且有較高的最佳化效能的群智能最佳化演算法。在粒子群最佳化演算法中，將需要設置慣性速度權重、個體最佳解權重、群體最佳解權重等多個超參數值。然而，目前在粒子群最佳化演算法相關研究的超參數值設定多為根據經驗或試誤的方式來求解適合的超參數值。但這樣的方式，可能不一定能找到超參數最佳解，並且可能需要較多的計算時間。

有鑑於此，本研究在既有的研究基礎[7]上提出一個分析框架，可以分析粒子群最佳化演算法在各個不同的適應函數中的平均適應函數值，並且從最佳化平均適應函數值來決定最合適的超參數值。並且深入探討超參數最佳解與其他變量和其他超參數之間的關聯，以及歸納出粒子群最佳化演算法超參數設置原則。

本論文共分為五個章節。第 2 節中介紹粒子群最佳化演算法，以及說明粒子群最佳化演算法的超參數。第 3 節提出本研究的分析框架，並且運用分析框架深入討論粒子群最佳化演算法在三個不同的適應函數的超參數最佳解，以及在 3.5 節中歸納本研究在粒子群最佳化演算法超參數設置的發現和建議。第 4 節將提供實例驗證本研究的超參數最佳解在各個適應函數的收斂表現。第 5 節總結本研究貢獻和討論未來研究方向。

## 2. 粒子群最佳化演算法

粒子群最佳化演算法由學者 James Kennedy 和 Russell Eberhart 在 1995 年[8]提出，由學者 Yuhui Shi 和 Russell Eberhart 在 1998 年[9]修改和完善最佳化演算法，並且被廣泛應用在各個領域。其中，粒子群最佳化演算法在最佳化的過程中，主要根據適應函數值作為最佳化目標，分析慣性速度為 $v_w$、粒子變量 $w$ 值與目前個體最佳解的距離為 $o_{p,w}$、粒子變量 $w$ 值與目前群體最佳解的距離為 $o_{g,w}$ 來最

佳化變量 $w$ 值，如公式(1)所示[8]-[9]。因此，需要設定超參數慣性速度權重$\alpha$、個體最佳解權重$c_1$、群體最佳解權重$c_2$，用以調整變量 $w$ 值，如公式(2)所示[8]-[9]。其中，$r_{1,w}$和$r_{2,w}$為隨機數。

$$v_{w,new} = \alpha v_w + r_{1,w}c_1 o_{p,w} + r_{2,w}c_2 o_{g,w} \quad (1)$$

$$\begin{aligned}w_{new} &= w + v_{w,new} \\ &= w + \alpha v_w + r_{1,w}c_1 o_{p,w} + r_{2,w}c_2 o_{g,w}\end{aligned} \quad (2)$$

## 3. 研究方法

本研究為證明粒子群最佳化演算法的超參數最佳解，提出一個分析框架，通過分析框架計算平均適應函數值，並且根據最佳化平均適應函數值來推導超參數的最佳解。此方法可以應用在最大化或最小化平均適應函數值，在本研究中以最小化為例進行說明。在 3.1 節中說明分析框架的具體步驟和流程，並且在 3.2 節和 3.3 節分別以單變量的適應函數和多變量的適應函數來說明如何最佳化超參數。在 3.4 節以線性迴歸的適應函數來分析超參數最佳解。最後，在 3.5 節小結各個適應函數推導過程的發現及其差異。

### 3.1 分析框架

圖 1 是本研究提出的分析框架其具體流程。先定義適應函數，確定求取適應函數的最大值或最小值，並假設適應函數中的變量呈均勻分佈時推導平均適應函數值。例如，以粒子群最佳化演算法對神經網路模型進行最佳化時，神經網路模型中的變量 $w$ 和變量 $b$ 等在初始值時主要採用隨機數，所以在實驗多次後將呈現均勻分佈。因此，在上述的假設，根據粒子群最佳化演算法推導在進行最佳化後的平均適應函數值。最後，再根據此平均適應函數值進行最佳化，推導在平均適應函數值最佳化時的超參數最佳解。

### 3.2 適應函數 1—單變量分析

為演示如何運用分析框架證明粒子群最佳化演算法超參數最佳解，在本節中以單變量的適應函數為例進行超參數最佳化。

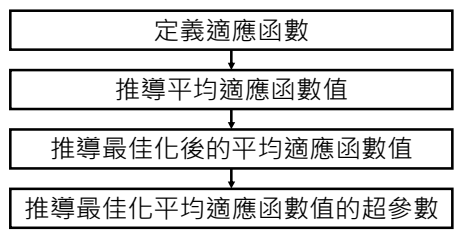

圖 1  分析框架

#### 3.2.1 函數 1 定義

**函數 1** 是單變量的適應函數$F_1(w)$如公式(3)所示，包含單變量 $w$。在此例中，以$F_1(w)$值作為適應函數值，並且最佳化目標是最小化適應函數值(即$F_1(w)$值)。假設變量 $w$ 服從均勻分佈，並且介於 0 到 1 之間，則可以通過公式(4)估計平均適應函數值。

$$F_1(w) = (w - 0.5)^2 \quad (3)$$

$$\int F_1(w)\,dw = \int (w - 0.5)^2\,dw \quad (4)$$

#### 3.2.2 函數 1 的超參數最佳化

在本節中將採用粒子群最佳化演算法對**函數 1** 進行最佳化，並且討論的粒子群最佳化演算法超參數的最佳解。其中，變量 $w$ 的最佳化主要通過超參數慣性速度權重$\alpha$、個體最佳解權重$c_1$、群體最佳解權重$c_2$ 來修正。因此，將修正後的值代入**函數 1** 中來計算經過粒子群最佳化演算法最佳化計算後的平均適應函數值，其中慣性速度為$v_w$、與個體最佳解的距離為$o_{p,w}$、與群體最佳解的距離為$o_{g,w}$、$r_{1,w}$和$r_{2,w}$為隨機數，如公式(5)所示。

為最佳化超參數(即慣性速度權重$\alpha$、個體最佳解權重$c_1$、群體最佳解權重$c_2$)和最小化平均適應函數值，定義函數$G_1(\alpha, c_1, c_2)$，如公式(6)所示。並且對公式(6)的$\alpha$微分求一階導函數為 0 時的$\alpha$值(即慣性速度權重最佳解$\alpha^*$)，如公式(7)所示；同理，可以推導個體最佳解權重$c_1$和群體最佳解權重來修

$$\int F_1(w + \alpha v_w + r_{1,w}c_1 o_{p,w} + r_{2,w}c_2 o_{g,w})\, dw$$

$$= \int (w + \alpha v_w + r_{1,w}c_1 o_{p,w} + r_{2,w}c_2 o_{g,w} - 0.5)^2\, dw \qquad (5)$$

$$= \frac{(w + \alpha v_w + r_{1,w}c_1 o_{p,w} + r_{2,w}c_2 o_{g,w} - 0.5)^3}{3}$$

$$G_1(\alpha, c_1, c_2) = \int F_1(w + \alpha v_w + r_{1,w}c_1 o_{p,w} + r_{2,w}c_2 o_{g,w})\, dw \qquad (6)$$

$$\alpha^* = \underset{0 \le \alpha \le 1}{arg\ min}\, G_1(\alpha, c_1, c_2) = \underset{0 \le \alpha \le 1}{arg\ min}\, \frac{(w + \alpha v_w + r_{1,w}c_1 o_{p,w} + r_{2,w}c_2 o_{g,w} - 0.5)^3}{3}$$

$$\Rightarrow \alpha^* = \frac{(w - 0.5) + (r_{1,w}c_1 o_{p,w} + r_{2,w}c_2 o_{g,w})}{-v_w} = \frac{\delta_1 + (r_{1,w}c_1 o_{p,w} + r_{2,w}c_2 o_{g,w})}{-v_w} \qquad (7)$$

$$c_1^* = \underset{0 \le c_1 \le 1}{arg\ min}\, G_1(\alpha, c_1, c_2) = \underset{0 \le c_1 \le 1}{arg\ min}\, \frac{(w + \alpha v_w + r_{1,w}c_1 o_{p,w} + r_{2,w}c_2 o_{g,w} - 0.5)^3}{3}$$

$$\Rightarrow c_1^* = \frac{(w - 0.5) + (\alpha v_w + r_{2,w}c_2 o_{g,w})}{-r_{1,w}o_{p,w}} = \frac{\delta_1 + (\alpha v_w + r_{2,w}c_2 o_{g,w})}{-r_{1,w}o_{p,w}} \qquad (8)$$

$$c_2^* = \underset{0 \le c_2 \le 1}{arg\ min}\, G_1(\alpha, c_1, c_2) = \underset{0 \le c_2 \le 1}{arg\ min}\, \frac{(w + \alpha v_w + r_{1,w}c_1 o_{p,w} + r_{2,w}c_2 o_{g,w} - 0.5)^3}{3}$$

$$\Rightarrow c_2^* = \frac{(w - 0.5) + (\alpha v_w + r_{1,w}c_1 o_{p,w})}{-r_{2,w}o_{g,w}} = \frac{\delta_1 + (\alpha v_w + r_{1,w}c_1 o_{p,w})}{-r_{2,w}o_{g,w}} \qquad (9)$$

正$c_2$，如公式(8)和公式(9)所示。其中，如果把$w - 0.5$定義為函數 1 的誤差$\delta_1$，則可以觀察到超參數作為變量的加權來調整，令誤差最小化。

### 3.3 適應函數 2—多變量分析

為演示如何運用分析框架證明粒子群最佳化演算法在多變量適應函數的超參數最佳解，在本節中以雙變量的適應函數為例進行超參數最佳化。

### 3.3.1 函數 2 定義

函數 2 是雙變量的適應函數$F_2(w, b)$如公式(10)所示，包含變量 $w$ 和變量 $b$。在此例中，以$F_2(w, b)$值作為適應函數值，並且最佳化目標是最小化適應函數值(即$F_2(w, b)$值)。假設變量 $w$ 和變量 $b$ 服從均勻分佈，並且介於 0 到 1 之間，則可以通過公式

(11)估計平均適應函數值。

$$F_2(w, b) = (w + b)^2 \qquad (10)$$

$$\iint F_2(w, b)\, dw\, db = \iint (w + b)^2\, dw\, db \qquad (11)$$

### 3.3.2 函數 2 的超參數最佳化

在本節中將採用粒子群最佳化演算法對函數 2 進行最佳化，並且討論的粒子群最佳化演算法超參數的最佳解。其中，變量 $w$ 和變量 $b$ 的最佳化主要通過超參數慣性速度權重$\alpha$、個體最佳解權重$c_1$、群體最佳解權重$c_2$來修正。因此，將修正後的值代入函數 2 中來計算經過粒子群最佳化演算法最佳化計算後的平均適應函數值，其中，變量 $w$ 慣

性速度為$v_w$、變量 $b$ 慣性速度為$v_b$、變量 $w$ 與個體最佳解的距離為$o_{p,w}$、變量 $w$ 與群體最佳解的距離為$o_{g,w}$、變量 $b$ 與個體最佳解的距離為$o_{p,b}$、變量 $b$ 與群體最佳解的距離為$o_{g,b}$、以及$r_{1,w}$、$r_{2,w}$、$r_{1,b}$、$r_{2,b}$為隨機數,如公式(12)所示。

為最佳化超參數(即慣性速度權重$\alpha$、個體最佳解權重$c_1$、群體最佳解權重$c_2$)和最小化平均適應函數值,定義函數$G_2(\alpha,c_1,c_2)$,如公式(13)所示。並且對公式(13)的$\alpha$微分求一階導函數為 0 時的$\alpha$值(即慣性速度權重最佳解$\alpha^*$),如公式(14)所示;同理,可以推導個體最佳解權重$c_1$和群體最佳解權重來修正$c_2$,如公式(15)和公式(16)所示。其中,如果

$$
\begin{aligned}
&\iint F_2(w+\alpha v_w+r_{1,w}c_1 o_{p,w}+r_{2,w}c_2 o_{g,w}, b+\alpha v_b+r_{1,b}c_1 o_{p,b}+r_{2,b}c_2 o_{g,b})\,dw\,db \\
&=\iint \left((w+\alpha v_w+r_{1,w}c_1 o_{p,w}+r_{2,w}c_2 o_{g,w})+(b+\alpha v_b+r_{1,b}c_1 o_{p,b}+r_{2,b}c_2 o_{g,b})\right)^2 dw\,db \\
&=\frac{\left((w+\alpha v_w+r_{1,w}c_1 o_{p,w}+r_{2,w}c_2 o_{g,w})+(b+\alpha v_b+r_{1,b}c_1 o_{p,b}+r_{2,b}c_2 o_{g,b})\right)^{12}}{12}
\end{aligned}
\quad (12)
$$

$$
G_2(\alpha,c_1,c_2)=\iint F_2\big(w+\alpha v_w+r_{1,w}c_1 o_{p,w}+r_{2,w}c_2 o_{g,w},\, b+\alpha v_b+r_{1,b}c_1 o_{p,b}+r_{2,b}c_2 o_{g,b}\big)\,dw\,db \quad (13)
$$

$$
\begin{aligned}
\alpha^* &= \arg\min_{0\le\alpha\le 1} G_2(\alpha,c_1,c_2) \\
&= \arg\min_{0\le\alpha\le 1} \frac{\left((w+\alpha v_w+r_{1,w}c_1 o_{p,w}+r_{2,w}c_2 o_{g,w})+(b+\alpha v_b+r_{1,b}c_1 o_{p,b}+r_{2,b}c_2 o_{g,b})\right)^{12}}{12} \\
\Rightarrow \alpha^* &= \frac{(w+b)+[c_1(r_{1,w}o_{p,w}+r_{1,b}o_{p,b})+c_2(r_{2,w}o_{p,w}+r_{2,b}o_{g,b})]}{-(v_w+v_b)} \\
&= \frac{\delta_2+[c_1(r_{1,w}o_{p,w}+r_{1,b}o_{p,b})+c_2(r_{2,w}o_{p,w}+r_{2,b}o_{g,b})]}{-(v_w+v_b)}
\end{aligned}
\quad (14)
$$

$$
\begin{aligned}
c_1^* &= \arg\min_{0\le c_1\le 1} G_2(\alpha,c_1,c_2) \\
&= \arg\min_{0\le c_1\le 1} \frac{\left((w+\alpha v_w+r_{1,w}c_1 o_{p,w}+r_{2,w}c_2 o_{g,w})+(b+\alpha v_b+r_{1,b}c_1 o_{p,b}+r_{2,b}c_2 o_{g,b})\right)^{12}}{12} \\
\Rightarrow c_1^* &= \frac{(w+b)+[\alpha(v_w+v_b)+c_2(r_{2,w}o_{g,w}+r_{2,b}o_{g,b})]}{-(r_{1,w}o_{p,w}+r_{1,b}o_{p,b})} \\
&= \frac{\delta_2+[\alpha(v_w+v_b)+c_2(r_{2,w}o_{g,w}+r_{2,b}o_{g,b})]}{-(r_{1,w}o_{p,w}+r_{1,b}o_{p,b})}
\end{aligned}
\quad (15)
$$

$$
\begin{aligned}
c_2^* &= \arg\min_{0\le c_2\le 1} G_2(\alpha,c_1,c_2) \\
&= \arg\min_{0\le c_2\le 1} \frac{\left((w+\alpha v_w+r_{1,w}c_1 o_{p,w}+r_{2,w}c_2 o_{g,w})+(b+\alpha v_b+r_{1,b}c_1 o_{p,b}+r_{2,b}c_2 o_{g,b})\right)^{12}}{12} \\
\Rightarrow c_2^* &= \frac{(w+b)+[\alpha(v_w+v_b)+c_1(r_{1,w}o_{p,w}+r_{1,b}o_{p,b})]}{-(r_{2,w}o_{g,w}+r_{2,b}o_{g,b})} \\
&= \frac{\delta_2+[\alpha(v_w+v_b)+c_1(r_{1,w}o_{p,w}+r_{1,b}o_{p,b})]}{-(r_{2,w}o_{g,w}+r_{2,b}o_{g,b})}
\end{aligned}
\quad (16)
$$

把 $w + b$ 定義為函數 2 的誤差 $\delta_2$，則可以觀察到超參數作為變量的加權來調整，令誤差最小化。

### 3.4 適應函數 3—迴歸應用

為演示如何運用分析框架證明粒子群最佳化演算法在迴歸及神經網路模型適應函數的超參數最佳解，本節中以線性迴歸的適應函數為例進行超參數最佳化。

#### 3.4.1 函數 3 定義

函數 3 是雙變量的適應函數 $F_3(w, b)$ 如公式(17)所示，包含變量 $w$ 和變量 $b$。在此例中，以 $F_3(w, b)$ 值作為適應函數值，並且最佳化目標是最小化適應函數值（即 $F_3(w, b)$ 值）。假設變量 $w$ 和變量 $b$ 服從均勻分佈，並且介於 0 到 1 之間，則可以通過公式(18)估計平均適應函數值。

$$F_3(w, b) = (wx + b - y)^2 \tag{17}$$

$$\iint F_3(w, b)\, dw\, db = \iint (wx + b - y)^2\, dw\, db \tag{18}$$

#### 3.4.2 函數 3 的超參數最佳化

在本節中將採用粒子群最佳化演算法對**函數 3** 進行最佳化，並且討論的粒子群最佳化演算法超參數的最佳解。其中，變量 $w$ 和變量 $b$ 的最佳化主要通過超參數慣性速度權重 $\alpha$、個體最佳解權重 $c_1$、群體最佳解權重 $c_2$ 來修正。因此，將修正後的值代入**函數 3** 中來計算經過粒子群最佳化演算法最佳化計算後的平均適應函數值，其中，變量 $w$ 慣性速度為 $v_w$、變量 $b$ 慣性速度為 $v_b$、變量 $w$ 與個體最佳解的距離為 $o_{p,w}$、變量 $w$ 與群體最佳解的距離為 $o_{g,w}$、變量 $b$ 與個體最佳解的距離為 $o_{p,b}$、變量 $b$ 與群體最佳解的距離為 $o_{g,b}$、以及 $r_{1,w}$、$r_{2,w}$、$r_{1,b}$、$r_{2,b}$ 為隨機數，如公式(19)所示。

為最佳化超參數（即慣性速度權重 $\alpha$、個體最佳解權重 $c_1$、群體最佳解權重 $c_2$）和最小化平均適應函數值，定義函數 $G_3(\alpha, c_1, c_2)$，如公式(20)所示。並且對公式(20)的 $\alpha$ 微分求一階導函數為 0 時的 $\alpha$ 值（即慣性速度權重最佳解 $\alpha^*$），如公式(21)所示；同理，可以推導個體最佳解權重 $c_1$ 和群體最佳解權重來修正 $c_2$，如公式(22)和公式(23)所示。其中，如果把 $wx + b - y$ 定義為函數 3 的誤差 $\delta_3$，則可以觀察到超參數作為變量的加權來調整，令誤差最小化。

### 3.5 小結與討論

本文定義 3 個適應函數，分別從單變量、多變量、以及迴歸應用，採用最小平方誤差作為目標來觀察超參數最佳解。通過公式推導結果，可以歸納出下面幾個設置原則，未來可推廣到神經網路模型最佳化等應用。

1. 由公式(7)、公式(8)、公式(9)、公式(14)、

$$\iint F_3(w + \alpha v_w + r_{1,w} c_1 o_{p,w} + r_{2,w} c_2 o_{g,w}, b + \alpha v_b + r_{1,b} c_1 o_{p,b} + r_{2,b} c_2 o_{g,b})\, dw\, db$$

$$= \iint \left((w + \alpha v_w + r_{1,w} c_1 o_{p,w} + r_{2,w} c_2 o_{g,w})x + (b + \alpha v_b + r_{1,b} c_1 o_{p,b} + r_{2,b} c_2 o_{g,b}) - y\right)^2 dw\, db \tag{19}$$

$$= \frac{\left((w + \alpha v_w + r_{1,w} c_1 o_{p,w} + r_{2,w} c_2 o_{g,w})x + (b + \alpha v_b + r_{1,b} c_1 o_{p,b} + r_{2,b} c_2 o_{g,b}) - y\right)^{12}}{12}$$

$$G_3(\alpha, c_1, c_2) = \iint F_3(w + \alpha v_w + r_{1,w} c_1 o_{p,w} + r_{2,w} c_2 o_{g,w}, b + \alpha v_b + r_{1,b} c_1 o_{p,b} + r_{2,b} c_2 o_{g,b})\, dw\, db \tag{20}$$

$$\alpha^* = \underset{0\le\alpha\le1}{arg\ min}\, G_3(\alpha, c_1, c_2)$$

$$= \underset{0\le\alpha\le1}{\arg\min} \frac{\left((w+\alpha v_w + r_{1,w}c_1 o_{p,w} + r_{2,w}c_2 o_{g,w})x + (b+\alpha v_b + r_{1,b}c_1 o_{p,b} + r_{2,b}c_2 o_{g,b}) - y\right)^{12}}{12}$$

$$\Rightarrow \alpha^* = \frac{(wx+b-y) + (r_{1,w}c_1 o_{p,w}x + r_{2,w}c_2 o_{g,w}x + r_{1,b}c_1 o_{p,b} + r_{2,b}c_2 o_{g,b})}{-(v_w x + v_b)}$$

$$= \frac{\delta_3 + \left[c_1(r_{1,w}o_{p,w}x + r_{1,b}o_{p,b}) + c_2(r_{2,w}o_{g,w}x + r_{2,b}o_{g,b})\right]}{-(v_w x + v_b)} \quad (21)$$

$$c_1^* = \underset{0\le c_1\le1}{arg\ min}\, G_3(\alpha, c_1, c_2)$$

$$= \underset{0\le c_1\le1}{\arg\min} \frac{\left((w+\alpha v_w + r_{1,w}c_1 o_{p,w} + r_{2,w}c_2 o_{g,w})x + (b+\alpha v_b + r_{1,b}c_1 o_{p,b} + r_{2,b}c_2 o_{g,b}) - y\right)^{12}}{12}$$

$$\Rightarrow c_1^* = \frac{(wx+b-y) + (\alpha v_w x + r_{2,w}c_2 o_{g,w}x + \alpha v_b + r_{2,b}c_2 o_{g,b})}{-(r_{1,w}o_{p,w}x + r_{1,b}o_{p,b})}$$

$$= \frac{\delta_3 + \left[\alpha(v_w x + v_b) + c_2(r_{2,w}o_{g,w}x + r_{2,b}o_{g,b})\right]}{-(r_{1,w}o_{p,w}x + r_{1,b}o_{p,b})} \quad (22)$$

$$c_2^* = \underset{0\le c_2\le1}{arg\ min}\, G_3(\alpha, c_1, c_2)$$

$$= \underset{0\le c_2\le1}{\arg\min} \frac{\left((w+\alpha v_w + r_{1,w}c_1 o_{p,w} + r_{2,w}c_2 o_{g,w})x + (b+\alpha v_b + r_{1,b}c_1 o_{p,b} + r_{2,b}c_2 o_{g,b}) - y\right)^{12}}{12}$$

$$\Rightarrow c_2^* = \frac{(wx+b-y) + (\alpha v_w x + r_{1,w}c_1 o_{p,w}x + \alpha v_b + r_{1,b}c_1 o_{p,b})}{-(r_{2,w}o_{g,w}x + r_{2,b}o_{g,b})}$$

$$= \frac{\delta_3 + \left[\alpha(v_w x + v_b) + c_2(r_{1,w}o_{p,w}x + r_{1,b}o_{p,b})\right]}{-(r_{2,w}o_{g,w}x + r_{2,b}o_{g,b})} \quad (23)$$

公式(15)、公式(16)、公式(21)、公式(22)、公式(23)可以觀察到當誤差(即$\delta_1$、$\delta_2$、$\delta_3$)越大時，超參數慣性速度權重$\alpha$、個體最佳解權重$c_1$、群體最佳解權重$c_2$都適合採用更大值，以快速降低誤差。因此，在訓練初期，建議可以採用較大的超參數值來降低誤差。

2. 公式(7)、公式(14)、公式(21)中的分母負號主要為修正方向，可以絕對值來觀察修正幅度。當個體最佳解權重$c_1$、群體最佳解權重$c_2$的係數值絕對值總和大於慣性速度權重$\alpha$的係數值絕對值總和時(如：$|c_1(r_{1,w}o_{p,w}x + r_{1,b}o_{p,b}) + c_2(r_{2,w}o_{p,w}x + r_{2,b}o_{p,b})| > |-(v_w x + v_b)|$)，則需要加大慣性速度權重$\alpha$以維持慣性速度的影響力。其中，當輸入變量數(即$x$的數量)增加時，由於分子和分母都有存在輸入變量，所以主要影響仍是慣性速度權重$\alpha$、個體最佳解權重$c_1$、群體最佳解權重$c_2$的係數值，以係數值決定最佳慣性速度權重$\alpha^*$。

3. 公式(8)、公式(15)、公式(22)中的分母負號主要為修正方向，可以絕對值來觀察修正幅度。當慣性速度權重$\alpha$、群體最佳解權重$c_2$的係數值絕對值總和大於個體最佳解權重$c_1$的係數值絕對值總和時(如：$|\alpha(v_w x + v_b) + c_2(r_{2,w}o_{p,w}x + r_{2,b}o_{p,b})| > |-(r_{1,w}o_{p,w}x + r_{1,b}o_{p,b})|$)，則需要加大個體最佳解權重$c_1$以維持個體最佳解的影響力。其中，當輸入變量數(即$x$的數量)增加時，由於分子和分母都有存

在輸入變量,所以主要影響仍是慣性速度權重$\alpha$、個體最佳解權重$c_1$、群體最佳解權重$c_2$的係數值,以係數值決定最佳個體最佳解權重$c_1^*$。

4. 公式(9)、公式(16)、公式(23)中的分母負號主要為修正方向,可以絕對值來觀察修正幅度。當慣性速度權重$\alpha$、個體最佳解權重$c_1$的係數值絕對值總和大於群體最佳解權重$c_2$的係數值絕對值總和時(如:$|\alpha(v_w x + v_b) + c_2(r_{1,w} o_{p,w} x + r_{1,b} o_{p,b})| > |-(r_{2,w} o_{g,w} x + r_{2,b} o_{g,b})|$),則需要加大群體最佳解權重$c_2$以維持群體最佳解的影響力。其中,當輸入變量數(即$x$的數量)增加時,由於分子和分母都有存在輸入變量,所以主要影響仍是慣性速度權重$\alpha$、個體最佳解權重$c_1$、群體最佳解權重$c_2$的係數值,以係數值決定最佳群體最佳解權重$c_2^*$。

## 4. 實驗分析與討論

為驗證本研究提出的方法效率和最佳化結果,把本研究方法與預設超參數值的結果分別在不同的適應函數比較收斂結果。其中,在本節中的超參數初始值和變量設定如下:粒子數量為2、慣性速度權重$\alpha = 0.5$、個體最佳解權重$c_1 = 1$、群體最佳解權重$c_2 = 1$、以及函數3中的輸入變量$x = 0.5$且輸出變量$y = 0.1 \times x + 0.2$。

根據上述超參數初始值和變量設定下,在不同的適應函數收斂結果如圖2、圖3、圖4所示。由實驗結果可以觀察到,本研究提出的方法之超參數最佳解,即使在僅有2個粒子數的情況下,對於每個適應函數都可以在20個Epoch內收斂到接近零誤差。如果超參數採用固定權重值,在僅有2個粒子數的情況下,將可能無法收斂到較低的誤差。因此,本研究方法的超參數相較於採用預設超參數或隨機產生的超參數,將可以更快收斂到最小誤差(即最小損失)。除此之外,本研究方法未來也可以求解不同的適應函數,並且也適用於求解最大值。

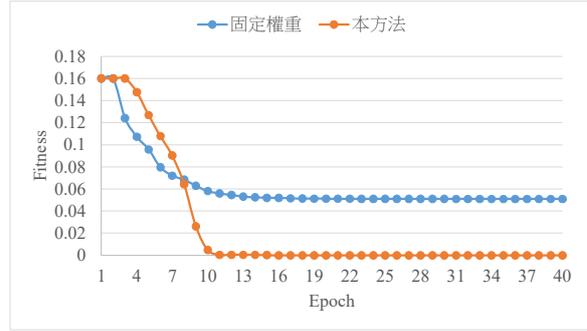

圖2 函數1收斂結果

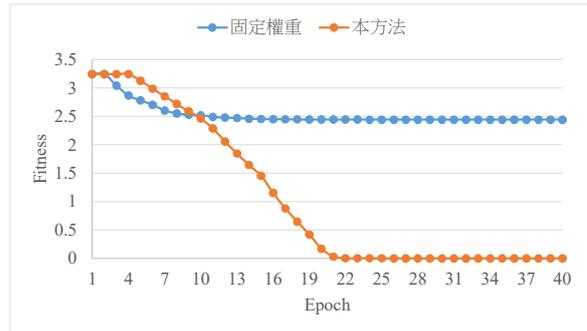

圖3 函數2收斂結果

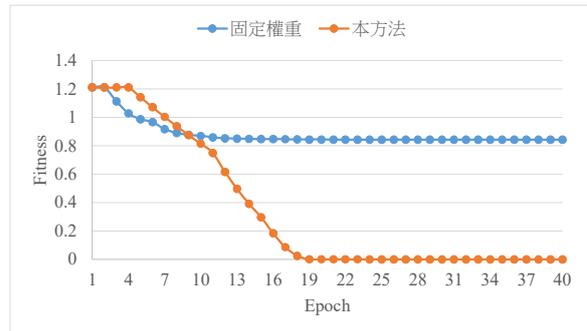

圖4 函數3收斂結果

## 5. 結論與未來研究

本研究為提出一個分析框架,分別分析在三個適應函數的平均適應函數值,並以最小化平均適應函數值來取得每個超參數的最佳解。並且,通過實驗證明,本研究提出的方法得到較快且較好的收斂結果。此外,在3.5節中歸納和總結粒子群最佳化演算法超參數在不同適應函數的變化和設置原則,可以作為未來其他模型設置超參數時參考。未來可以修改適應函數,應用到深度學習模型最佳化。

## 參考文獻

# How to Prove the Optimized Values of Hyperparameters for Particle Swarm Optimization?

Abel C. H. Chen

Chunghwa Telecom Co., Ltd.

chchen.scholar@gmail.com

ORCID 0000-0003-3628-3033

**Abstract**

In recent years, several swarm intelligence optimization algorithms have been proposed to be applied for solving a variety of optimization problems. However, the values of several hyperparameters should be determined. For instance, although Particle Swarm Optimization (PSO) has been applied for several applications with higher optimization performance, the weights of inertial velocity, the particle's best known position and the swarm's best known position should be determined. Therefore, this study proposes an analytic framework to analyze the optimized average-fitness-function-value (AFFV) based on mathematical models for a variety of fitness functions. Furthermore, the optimized hyperparameter values could be determined with a lower AFFV for minimum cases. Experimental results show that the hyperparameter values from the proposed method can obtain higher efficiency convergences and lower AFFVs.

***Keywords***: *Particle swarm optimization, hyperparameter, optimized value*